\documentclass[10pt,twocolumn,letterpaper]{article}

\usepackage[pagenumbers]{cvpr} %

\input{preamble}

\definecolor{cvprblue}{rgb}{0.21,0.49,0.74}
\usepackage[pagebackref,breaklinks,colorlinks,allcolors=cvprblue]{hyperref}

\newcommand{\appendixmode}{}
\newcommand{\mainpaper}{main text}
\newcommand{\suppmaterial}{appendix}

\def\confName{CVPR}
\def\confYear{2027}

\title{Efficient Video Retrieval Augmented Generation via Token Allocation}

\author{Ayush Barik$^{1}$ \qquad Fan Lai$^{1\dagger}$ \qquad Francis Y. Yan$^{1\dagger}$\\
$^{1}$University of Illinois Urbana-Champaign\\
{\tt\small \{barik2, fanlai, fyy\}@illinois.edu}}

\begin{document}
\maketitle
\iftoggle{cvprfinal}{\begin{NoHyper}\renewcommand{\thefootnote}{$\dagger$}\footnotetext{Equal advising contribution.}\end{NoHyper}}{}
\begin{abstract}
Efficient long-video understanding requires identifying relevant evidence while minimizing the tokens processed by vision-language models (VLMs). Existing video retrieval-augmented generation (RAG) methods rely on frame selection and token compression, but overlook how much each additional token is worth. We introduce \textbf{\name}, a plug-and-play framework that formulates video RAG as a token allocation problem over frames and subtitles under a shared budget. Our key insight is that the value of a token diminishes within each item and decreases further as similar evidence receives more tokens. \name captures both effects in a single objective and solves it with an efficient iterative water-filling algorithm that has a provable convergence guarantee. For multi-turn conversations, \name reuses the tokens processed in earlier turns and refines relevance with the VLM's attention at no extra cost. Our evaluations show that \name improves accuracy by up to 7.2 percentage points over state-of-the-art baselines at comparable compute. On multi-turn video understanding, \name achieves up to 1.5 percentage points higher accuracy with 1.7--17.3$\times$ fewer TFLOPs.
\end{abstract} 
    
\section{Introduction} \label{sec:intro}

Vision-language models (VLMs) have achieved remarkable progress in video understanding, enabling tasks ranging from action recognition to complex visual question answering~\cite{bai2025qwen3vl,wang2025internvl35}. However, extending these capabilities to long videos, such as sports broadcasts, personal recordings, and surveillance footage, remains challenging. Processing every frame can easily exceed the VLM's context window and incur prohibitive inference costs~\cite{chen2024fastv,yang2024visionzip}. Efficient long-video understanding requires identifying and processing the evidence most useful for answering each query.

\begin{figure}[t]
  \centering
  \includegraphics[width=0.9\linewidth]{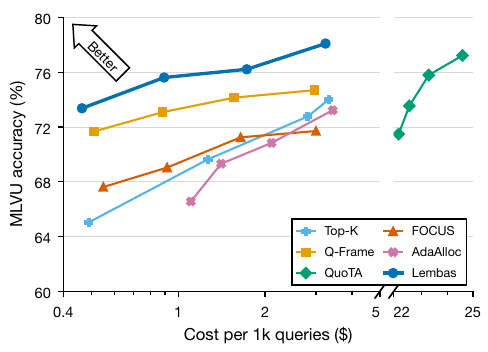}
  \vspace{-5pt}
  \caption{\name achieves the best accuracy-cost trade-off. Dollar cost is derived from OpenRouter API prices for Qwen3-VL-8B.}
  \vspace{-5pt}
  \label{fig:budget}
\end{figure}

Retrieval-augmented generation (RAG) addresses this challenge by retrieving relevant frames rather than processing the entire video. Embedding-based RAG retrieves frames based on semantic similarity to the query, while graph-based RAG connects entities and events to retrieve related evidence~\cite{luo2024videorag,ren2025videorag,shen2025vgent,yan2025ava}. However, these optimizations select relevant evidence items yet often overlook an inherent distinction: different queries require different amounts of evidence and visual detail. For example, identifying a small object may require fine-grained information from a particular frame, whereas reconstructing an event sequence may benefit from broader temporal coverage. 

Recent approaches improve this trade-off through adaptive frame selection and token compression. Frame selection methods identify relevant or diverse frames using retrieval heuristics~\cite{tang2025aks,zhu2026focus,ye2025tstar,peng2026qca,shi2026markey,wang2026dafs}, learned policies~\cite{yu2023sevila,yu2024framevoyager,buch2025flexible,lee2025refocus}, or agentic search~\cite{wang2024videoagent,zou2025air}. Token compression methods reduce visual redundancy through token merging or pruning~\cite{fu2025framefusion,tao2025dycoke,shao2025holitom}. More recent methods adapt token counts using relevance-proportional allocation~\cite{luo2025quota}, discrete resolution tiers~\cite{zhang2025qframe}, or agentic budget planning~\cite{an2026adaalloc}. Despite these advances, allocation is largely guided by independent relevance scores or predefined rules.

Our key insight is that \emph{token efficiency requires globally allocating tokens according to their marginal value}. We make two observations. First, additional tokens exhibit diminishing returns: once a frame preserves sufficient detail, allocating more tokens yields progressively smaller accuracy gains. Second, token value depends on other allocated evidence: a frame contributes less new information when similar frames already receive tokens. Naturally, one fundamental problem is to distribute a shared token budget across frames and subtitles to maximize their aggregate utility.

\begin{figure}[t]
  \centering
  \includegraphics[width=\linewidth]{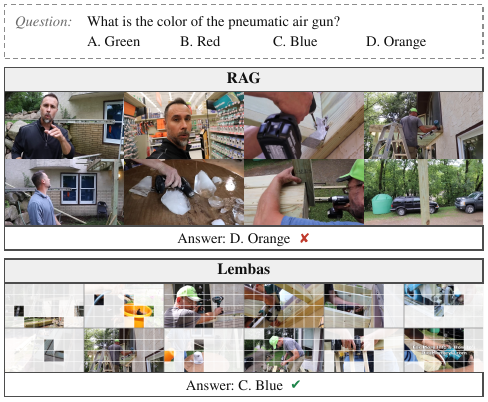}
  \caption{Varying the tokens per frame allows \name to retain evidence from more frames within a shared budget.}
  \label{fig:case}
\end{figure}

Building on these observations, we introduce \name, a plug-and-play framework for token-efficient video RAG (\cref{fig:overview}). Given frames and subtitles retrieved by an existing RAG system, \name captures their utility, allocates a shared token budget, and prunes each item accordingly before VLM inference. Realizing this design faces two challenges: modeling token value and optimizing allocations efficiently. Moreover, users rarely ask a single question about a video, and a conversation that spans multiple turns raises a third challenge: reusing evidence as allocations change across turns.

\para{Modeling token value.}
An item's contribution depends not only on its relevance to the query, but also on how many tokens it receives and how much of its content is already covered by other items. We characterize downstream accuracy across token counts and find that a concave, saturating rational function well captures diminishing returns (\cref{fig:value}). We combine this modeling with query relevance and redundancy awareness, and use attention-based pruning to realize the resulting token counts (\cref{sec:objective}).

\para{Solving token allocation.}
Accounting for redundancy makes token allocations interdependent, exploding the combination space. We develop iterative water-filling, which repeatedly updates diversity-aware relevance and solves a closed-form water-filling subproblem. Our solver converges to the global optimum of the continuous relaxation with an $O(1/\ell)$ objective gap after $\ell$ iterations (\cref{sec:allocation}).

\para{Reusing evidence across turns.} Multi-turn conversations often revisit the same evidence, but different queries may require different token counts, complicating KV-cache reuse. \name addresses this through query-independent token ranking, allowing subsequent turns to prefill only tokens missing from the cache. Moreover, we leverage the VLM's attention to cached evidence to refine its relevance for subsequent allocations without additional VLM calls (\cref{sec:multiturn}).

We evaluate \name on MLVU, LongVideoBench, and Video-MME with
Qwen3-VL-8B and InternVL3.5-4B (\cref{sec:exp}).
With comparable compute budgets (\cref{fig:budget,tab:main}), \name improves accuracy by 0.1--7.2 percentage points
over state-of-the-art baselines.
On the multi-turn benchmark MT-Video-Bench, \name achieves 0.3--1.5 percentage points higher accuracy than baselines
with 1.7--17.3$\times$ fewer TFLOPs.

 Overall, this paper makes the following contributions:
\begin{itemize}
  \item We introduce \name, a token-efficient video RAG framework that 
  formulates video understanding as a token allocation problem under a limited budget.
  \item We develop a Token Allocator that uses iterative water-filling to efficiently solve the token optimization problem.
  \item We extend \name to multi-turn conversations through incremental allocation and attention refinement, enabling evidence reuse as allocations change.
\end{itemize}

\section{Related Work}
\label{sec:related}

\paragraph{Retrieval-augmented generation for video understanding.}
Video RAG~\cite{lewis2020retrieval} extends retrieval-augmented generation to long videos, whose frames far exceed the context window of a VLM. Existing systems differ mainly in how they index a video and how they retrieve evidence from this index. For indexing, one line of work converts the video into textual surrogates, such as captions, speech transcripts, and on-screen text, and retrieves over them~\cite{ma2024drvideo,luo2024videorag}, while another embeds frames or clips directly and retrieves by query--visual similarity, including across video corpora~\cite{jeong2025videorag} and live streams~\cite{xie2026streamrag}. To capture relations that such flat indices miss, structured indices further organize the video into trees~\cite{wang2024videotree} or entity- and event-centric graphs~\cite{ren2025videorag,shen2025vgent,yan2025ava}. For retrieval, single-step similarity search has been extended to multi-step and agentic retrieval, which refines the retrieved evidence over several rounds~\cite{shen2025vgent,yan2025ava}. However, these systems improve \emph{what} to index and retrieve, and pass every retrieved item to the VLM at its full, uniform token cost. \name instead addresses \emph{how many} tokens each retrieved item should receive, operating as a plug-and-play module downstream of these retrievers.

\paragraph{Keyframe or token selection.} Rather than fully processing long videos to answer user questions, various methods intelligently decide \emph{which} frames to process~\cite{tang2025aks,zhu2026focus,li2026dig,liu2025bolt,sun2025mdp3,shi2026markey,peng2026qca,yu2024framevoyager,wang2026dafs}, either due to practical constraints (e.g., the context window~\cite{zhang2024longva}) or to improve efficiency via reduced token prefill costs~\cite{chen2024fastv,yang2024visionzip}. Token pruning has been widely studied in long-context LLM serving. H2O~\cite{h2o-neurips23} prunes tokens with smaller accumulated attention, while StreamingLLM~\cite{streamingllm-iclr24} retains the initial and recent tokens. Beyond \emph{which} frames or tokens to keep, recent and concurrent works decide \emph{how much} to spend on each frame, typically by routing frames to a few fixed resolution tiers through agentic search~\cite{an2026adaalloc} or hand-crafted rules~\cite{zhang2025qframe, luo2025quota}. \name instead treats this question as a resource allocation problem, formulating an objective, solving it with a convergence guarantee, and extending it to multi-turn conversations.

\paragraph{Conventional resource allocation.}
Resource allocation has been widely studied in networked and computing systems to distribute limited resources among competing demands~\cite{kelly1998rate,low1999optimization,ghodsi2011dominant}. Classical approaches include utility maximization and pricing mechanisms~\cite{palomar2006tutorial,chiang2007layering}, water-filling and power control~\cite{goldsmith1997capacity,yu2004iwf,scutari2008asynchronous}, and fairness-aware scheduling~\cite{isard2009quincy,hindman2011mesos,grandl2014multi}. \name brings this resource allocation perspective to video understanding, allocating a finite token resource across retrieved items via iterative water-filling.

\section{Method}
\label{sec:method}

\begin{figure*}[t]
  \centering
  \makebox[\linewidth][c]{\input{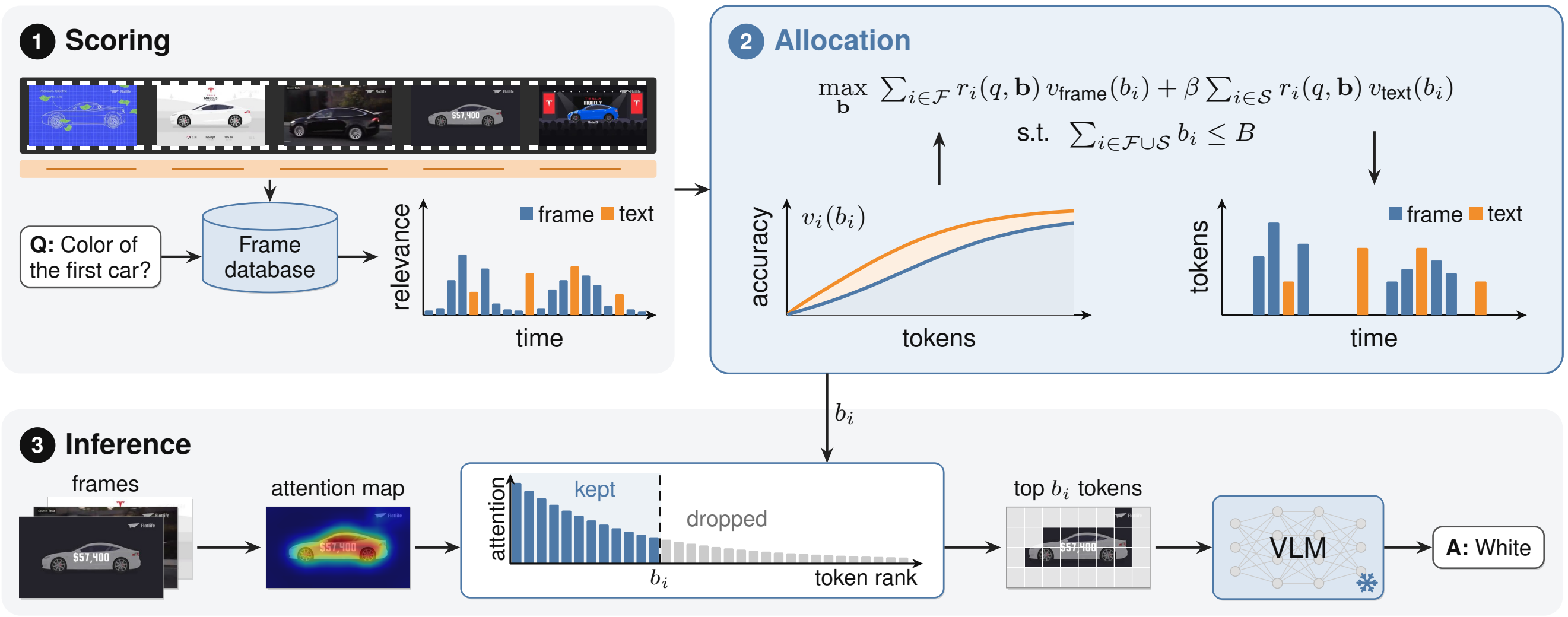}}
  \caption{An illustration of \name's workflow.}
  \label{fig:overview}
\end{figure*}

\name is a plug-and-play framework that makes video RAG token-efficient. As shown in \cref{fig:overview}, given a query, \name scores the relevance of each frame and subtitle, optimizes their allocation under a token budget, and retains the highest-ranked tokens from each selected item. Its Token Allocator exploits diminishing returns in token value and discounts redundant evidence. For multi-turn conversations, \name further reuses tokens already in the KV cache and refines their relevance using the VLM's attention.

\subsection{Problem Formulation}
\label{sec:overview}

Consider a video $V$ and query $q$. Conventional Top-$K$ video RAG retrieves $K$ frames with the highest relevance to $q$. Let $f_1,\ldots,f_N$ be candidate frames and let $r_i(q)\geq 0$ denote the retrieval relevance of frame $i$. Viewing it from an optimization lens, this is an integer linear program:
\begin{equation}
  \max_{\mathbf{x}\in\{0,1\}^N}
  \sum_{i=1}^{N}r_i(q)x_i
  \quad\text{s.t.}\quad \sum_{i=1}^{N}x_i\leq K,
  \label{eq:topk}
\end{equation}
where $x_i$ indicates whether frame $i$ is selected. The optimum simply selects the $K$ highest-scoring frames. However, each selected frame is processed at a fixed token count. As a result, Top-$K$ cannot trade unnecessary detail in one frame for useful evidence from another.

\name aims to break this rigidity and achieve better efficiency by lifting the problem from frame selection to multimodal token allocation. Let $\mathcal F$ and $\mathcal S$ denote the candidate frames and subtitle segments, and let $b_i\in[0,b_i^{\max}]$ be the tokens allocated to item $i\in\mathcal F\cup\mathcal S$, where $b_i^{\max}$ is its full token count. The function $v_i(b_i)\in[0,1]$, which is $v_{\text{frame}}$ for a frame and $v_{\text{text}}$ for a subtitle, models the value retained when item $i$ is represented by $b_i$ tokens, and $r_i(q,\mathbf b)$ measures its relevance given both the query and the evidence allocated elsewhere. Frames and subtitles share a total budget $B$, with $\beta>0$ balancing their contributions. Then \name aims to optimize:
\begin{equation}
\begin{aligned}
  \max_{\mathbf{b}}\quad & \sum_{i \in \mathcal{F}} r_i(q, \mathbf{b})\, v_{\text{frame}}(b_i) + \beta \sum_{i \in \mathcal{S}} r_i(q, \mathbf{b})\, v_{\text{text}}(b_i) \\
  \text{s.t.}\quad & \sum_{i \in \mathcal{F} \cup \mathcal{S}} b_i \le B
\end{aligned}
  \label{eq:allocation}
\end{equation}
 However, this formulation faces three challenges in practice: \circnum{1}~how do we model the token value functions $v_{\text{frame}}$ and $v_{\text{text}}$ and the relevance $r_i$ of each item? (\cref{sec:objective}) \circnum{2}~how do we optimize the stated objective efficiently? (\cref{sec:allocation}) and \circnum{3}~how do we refine and reuse allocated tokens across multi-turn conversations? (\cref{sec:multiturn}).

\subsection{Constructing the Allocation Objective}
\label{sec:objective}

\paragraph{Token Value Function.}
Retrieval relevance estimates an item's relevance when it is fully represented, but not how much of that usefulness survives token pruning. In Eq.~\eqref{eq:allocation}, we model this using the token value functions $v_{\text{frame}}(b_i)$ and $v_{\text{text}}(b_i)$, which weigh an item's relevance by the fraction of its value preserved at token count $b_i$. To realize an arbitrary token count $b_i$, \name ranks the visual tokens of each frame once, offline and independently of the query, by the attention they receive in the last layer of the VLM's vision encoder, summed over all heads, and keeps the top $b_i$ tokens at their original positions, following the attention-based ranking of popular token pruners~\cite{chen2024fastv,yang2024visionzip}.

\begin{figure}[t]
  \centering
  \begin{minipage}[t]{0.485\linewidth}
    \centering
    \includegraphics[width=\linewidth]{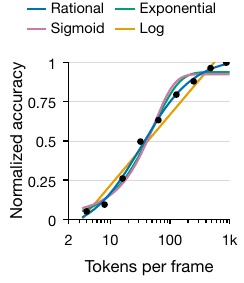}
    \caption{Candidate token-value functions fitted to the measured accuracy of frames (black dots).
    }
    \label{fig:value}
  \end{minipage}\hfill
  \begin{minipage}[t]{0.485\linewidth}
    \centering
    \includegraphics[width=\linewidth]{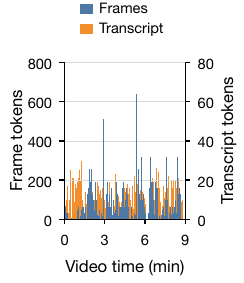}
    \caption{Example of \name's token allocation for a nine-minute video.
    }
    \label{fig:alloc}
  \end{minipage}
\end{figure}

To define the token value functions, we measure how Q\&A accuracy changes with the token count. On held-out validation sets of multiple video QA benchmarks~\cite{zhou2024mlvu,wu2024longvideobench,fu2024videomme}, we prune every frame of a short to medium-length video to $b_i$ tokens, feed the frames and the query to a VLM, and record the mean accuracy at each $b_i$. We then normalize the accuracies and fit several candidate functions (\cref{fig:value}). A simple rational function fits the measurements best, for both visual and textual evidence:
\begin{equation}
  v_{\text{frame}}(b_i)=\frac{b_i}{b_i+\tau_{\text{frame}}},\quad
  v_{\text{text}}(b_i)=\frac{b_i}{b_i+\tau_{\text{text}}},
  \label{eq:curve}
\end{equation}
where $\tau_{\rm frame}>0$ and $\tau_{\rm text}>0$ are modality-specific parameters estimated offline by least squares. The fits achieve $R^2>0.85$, which shows that a simple function suffices for this task over more complex methods (e.g., neural networks). Because a static function may not be robust to shifting query distributions, the parameters can be refitted during off-peak serving periods when query feedback (e.g., accuracy) is available.

\paragraph{Diversity-aware Relevance.}
The retrieval relevance $r_i(q)$ captures relevance but assumes that items are independent. However, in videos, nearby frames may depict nearly identical content; therefore, frames similar to already selected frames should be discounted. To capture this dependency, we discount each item's retrieval relevance $r_i(q)$ according to the token value already allocated to similar items. Let $s_{ij}$ denote the cosine similarity between item embeddings within the same modality. We define the complete allocation-dependent relevance as follows:
\begin{equation}
\begin{aligned}
  r_i(q,\mathbf b)&=r_i(q)-\eta\sum_{j\in\mathcal F}s_{ij}\,v_{\text{frame}}(b_j), & i&\in\mathcal F,\\
  r_i(q,\mathbf b)&=r_i(q)-\eta\sum_{j\in\mathcal S}s_{ij}\,v_{\text{text}}(b_j), & i&\in\mathcal S,
\end{aligned}
\label{eq:cond}
\end{equation}
where $\eta\geq0$ controls the redundancy penalty.

Substituting \cref{eq:curve,eq:cond} into \cref{eq:allocation} finally yields the full allocation objective which rewards query-relevant evidence, accounts for diminishing token value, and discourages redundant allocations.

\subsection{Efficient Token Allocation}
\label{sec:allocation}

Allocating discrete token counts directly involves many integer decisions that are coupled through the diversity term. Off-the-shelf integer solvers (e.g., Gurobi~\cite{gurobi}, CPLEX~\cite{cplex}) take exponential time in the worst case: on a sample hour-long video at 1 FPS, they run for more than a dozen seconds, exceeding the VLM's own inference time.

Hence, we optimize a continuous relaxation instead and round its solution to whole tokens afterward. The key is to change variables from token counts to \emph{token values} $z_i=v_i(b_i)$. Because our rational token value functions are concave and the similarity matrix is positive semidefinite, this relaxation turns Eq.~\eqref{eq:allocation} into a continuous convex program in the token values. To start, we solve a simpler subproblem in which the relevance is fixed, which naturally suggests a \emph{water-filling} algorithm, and then remove this restriction and extend it to \emph{iterative water-filling} for the full objective.

\paragraph{Water-Filling.}
Consider first the subproblem in which the relevance of each item is fixed to $\bar r_i$, its retrieval relevance $r_i(q)$ (scaled by $\beta$ for subtitles), so that it no longer depends on the allocation. The objective then reduces to a sum of concave terms $\bar r_i\,v_i(b_i)$ under a single budget, where each additional token of an item yields a smaller gain than the previous one. Hence, at the optimum, all partially allocated items must share the same marginal gain per token, which we denote by $\lambda$; otherwise, moving a token from an item with a lower marginal gain to one with a higher marginal gain would improve the objective. For our token value functions, the marginal gain of item $i$ is $\bar r_i\tau_i/(b_i+\tau_i)^2$, where $\tau_i$ is the token value parameter of its modality. Setting it to $\lambda$ and solving for $b_i$ yields
\begin{equation}
b_i=\left[\sqrt{\bar r_i\tau_i/\lambda}-\tau_i\right]_0^{b_i^{\max}},
\label{eq:waterfill}
\end{equation}
where $[\cdot]_0^{b_i^{\max}}$ clips the token count to $[0,b_i^{\max}]$. Intuitively, an item receives tokens only if its relevance exceeds $\lambda\tau_i$, and its token count then grows with the square root of its relevance rather than in proportion to it. A single scalar $\lambda$ thus decides both which items are selected and how many tokens each receives. Since the total allocation decreases monotonically with $\lambda$, a bisection finds the $\lambda$ that meets the budget $B$. This procedure is known as water-filling~\cite{boyd2004,yu2004iwf}.

\paragraph{Iterative Water-Filling.}
The full objective, however, discounts the relevance of each item by the token values of its similar items, so the relevance that water-filling requires depends on the very allocation it produces. We resolve this circular dependence using the Frank--Wolfe method~\cite{frank1956,jaggi2013}, which iteratively optimizes linear approximations of the objective. In terms of the token values, Eq.~\eqref{eq:allocation} reads $J(\mathbf z)=\bar{\mathbf r}^{\top}\mathbf z-\eta\,\mathbf z^{\top}\mathbf S\mathbf z$, where $\mathbf S$ is the block-diagonal matrix of within-modality similarities $s_{ij}$ (scaled by $\beta$ for subtitles). At iteration $\ell$, given the current token values $\bar{\mathbf z}$ (initialized to zero), we compute the diversity-discounted relevance as
\begin{equation}
\widetilde{\mathbf r}
=\nabla J(\bar{\mathbf z})
=\bar{\mathbf r}-2\eta\,\mathbf S\bar{\mathbf z}.
\label{eq:weights}
\end{equation}
Maximizing the linearized objective $\widetilde{\mathbf r}^{\top}\mathbf z$ under the token budget is exactly the fixed-relevance subproblem above. A single water-filling pass with $\widetilde r_i$ in place of $\bar r_i$ therefore yields its optimal allocation $\mathbf b^\star$, in which items with $\widetilde r_i\leq 0$ receive no tokens. However, directly adopting this allocation may overcorrect for redundancy, since it optimizes only the linearized objective. This motivates us to update the token values toward $\mathbf z^\star=\mathbf v(\mathbf b^\star)$ using a diminishing step size:
\begin{equation}
\bar{\mathbf z}\leftarrow
(1-\gamma_\ell)\bar{\mathbf z}
+\gamma_\ell\mathbf z^\star,
\qquad
\gamma_\ell=\frac{2}{\ell+1}.
\label{eq:average}
\end{equation}
The first iteration recovers standard water-filling, while subsequent iterations refine the allocation to account for redundant evidence. After $L$ iterations, we recover token counts as $b_i=v_i^{-1}(\bar z_i)$. The complete procedure is summarized in \cref{alg:iwf}. We next show that our solution achieves global convergence at a sublinear rate:

\begin{proposition}
\label{prop:convergence}
Let $J^\star$ be the global maximum of Eq.~\eqref{eq:allocation} over continuous token allocations, and let $J^{(\ell)}$ be the objective value after $\ell$ iterations of \cref{alg:iwf}. Then
\begin{equation}
  J^\star - J^{(\ell)} \le \frac{4\,\eta\,\sigma_{\max}D^2}{\ell + 2} = O\!\left(\frac{1}{\ell}\right),
\end{equation}
where $\sigma_{\max}$ is the largest eigenvalue of the frame and $\beta$-scaled subtitle similarity matrices and $D$ is the diameter of the feasible token-value set under budget $B$.
\end{proposition}

\noindent This bound follows from the standard Frank--Wolfe guarantee~\cite{jaggi2013}. The VLM consumes whole tokens, so we round the continuous allocation down, $\hat b_i=\lfloor b_i\rfloor$. Rounding keeps the allocation within the budget and removes less than one token from each item. Finally, \name retains the top $\hat b_i$ pre-ranked tokens for each item and presents the selected evidence in chronological order.

\paragraph{Optimality over Prior Allocation Rules.}
Prior methods allocate tokens by fixed rules rather than by optimizing an objective. QuoTA~\cite{luo2025quota} assigns each frame a token count proportional to its relevance, while Q-Frame~\cite{zhang2025qframe} assigns the same token count to all frames within a relevance tier. Neither rule is optimal, even when diversity is ignored. Because the token value functions are strictly concave, the water-filling allocation of Eq.~\eqref{eq:waterfill} is the unique maximizer of the fixed-relevance objective, and it differs from both rules. Specifically, its token counts grow with the square root of the relevance rather than in proportion to it, and items with $\bar r_i\le\lambda\tau_i$ receive no tokens at all. Proportional allocation, in contrast, keeps adding tokens to the most relevant frames after their value has saturated, while tiered allocation ignores the differences in relevance within a tier. Moreover, both rules score frames independently, whereas iterative water-filling also accounts for the redundancy among items and converges to the optimum of the full objective (\cref{prop:convergence}).

\begin{algorithm}[t]
\caption{Iterative Water-Filling}
\label{alg:iwf}
\small
\begin{algorithmic}[1]
\Require relevances $\bar r_i$, similarities $s_{ij}$, value parameters $\tau_i$, token limits $b_i^{\max}$, budget $B$, penalty $\eta$, iterations $L$, tolerance $\epsilon$
\Ensure token counts $\hat b_i$ with $\sum_i\hat b_i\le B$
\Statex Lines with index $i$ apply to every item $i\in\mathcal F\cup\mathcal S$.
\State Initialize token values $\bar z_i\gets 0$
\For{$\ell=1,\dots,L$}
  \State $\widetilde r_i\gets\bar r_i-2\eta\sum_{j}s_{ij}\,\bar z_j$ \Comment{discounted relevance, Eq.~\eqref{eq:weights}}
  \State $\lambda_{\rm lo}\gets 0,\;\;\lambda_{\rm hi}\gets\max_i\widetilde r_i/\tau_i$ \Comment{search range of $\lambda$}
  \State $b_i^\star\gets 0$ \Comment{best allocation within budget so far}
  \While{$\lambda_{\rm hi}-\lambda_{\rm lo}>\epsilon$} \Comment{water-filling by bisection}
    \State $\lambda\gets(\lambda_{\rm lo}+\lambda_{\rm hi})/2$
    \State $b_i\gets\big[\sqrt{\max(0,\widetilde r_i)\,\tau_i/\lambda}-\tau_i\big]_0^{b_i^{\max}}$ \Comment{Eq.~\eqref{eq:waterfill}}
    \If{$\sum_i b_i>B$}
      \State $\lambda_{\rm lo}\gets\lambda$ \Comment{over budget: raise $\lambda$}
    \Else
      \State $\lambda_{\rm hi}\gets\lambda,\;\;b_i^\star\gets b_i$ \Comment{within budget: lower $\lambda$}
    \EndIf
  \EndWhile
  \State $\gamma_\ell\gets 2/(\ell+1)$ \Comment{step size}
  \State $\bar z_i\gets(1-\gamma_\ell)\,\bar z_i+\gamma_\ell\,v_i(b_i^\star)$ \Comment{averaging, Eq.~\eqref{eq:average}}
\EndFor
\State $b_i\gets v_i^{-1}(\bar z_i)=\tau_i\bar z_i/(1-\bar z_i)$ \Comment{continuous token counts}
\State \Return $\hat b_i\gets\lfloor b_i\rfloor$ \Comment{round down to whole tokens}
\end{algorithmic}
\end{algorithm}

\subsection{Extending to Multi-Turn Conversations}
\label{sec:multiturn}

\begin{figure}[t]
  \centering
  \begin{minipage}[t]{0.47\linewidth}
    \centering
    \includegraphics[width=\linewidth]{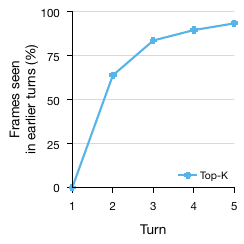}
    \caption{Later turns often reference frames already prefilled in prior turns.}
    \label{fig:reuse}
  \end{minipage}\hfill
  \begin{minipage}[t]{0.47\linewidth}
    \centering
    \includegraphics[width=\linewidth]{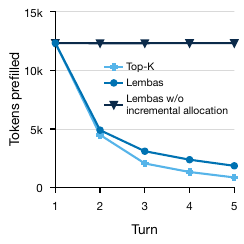}
    \caption{Incremental allocation allows \name to recover a large amount of the available reuse opportunities.}
    \label{fig:prefill}
  \end{minipage}
\end{figure}

In practical conversations, a user may ask several queries $q_1,\dots,q_T$ about the same video~\cite{hsieh2018focus}. Later queries often refer to evidence processed in earlier turns (\cref{fig:reuse}). Top-K can reuse a previously prefilled frame when the selection matches exactly, but token allocation is harder: the same item may receive different token counts at different turns. \name addresses this with \emph{incremental allocation}, which prefills only the difference between the current allocation and all previous ones, and \emph{attention refinement}, which refines the relevance of cached evidence with attention for free.

\paragraph{Incremental Allocation.}
Let $b_i^{(t)}$ be the token count requested for item $i$ at turn $t$, and let $c_i^{(t-1)}$ be the number of its highest-ranked tokens already in the KV cache, with $c_i^{(0)}=0$. The number of new tokens to prefill and the updated cache state are
\begin{equation}
\begin{aligned}
 p_i^{(t)}&=\max\!\left(0,b_i^{(t)}-c_i^{(t-1)}\right),\\
 c_i^{(t)}&=\max\!\left(c_i^{(t-1)},b_i^{(t)}\right).
\end{aligned}
\label{eq:incremental}
\end{equation}
As the KV cache is typically retained across turns in standard serving systems, evidence prefilled at an earlier turn remains available for later queries to leverage, and the budget $B$ applies to the evidence allocated at each turn. Because token rankings are fixed independently of the query, an item allocated more tokens at turn $t$ needs only the newly included ranks $c_i^{(t-1)}+1,\ldots,b_i^{(t)}$. \name appends these tokens to the existing context instead of inserting them into earlier positions, which would invalidate subsequent KV-cache entries. The model retains the applicable within-item visual or temporal positional information. As shown in \cref{fig:prefill}, incremental allocation thereby recovers a large share of the available reuse opportunities.

Since each token is prefilled at most once, summing \cref{eq:incremental} over turns bounds cumulative evidence tokens:
\begin{equation}
 \sum_{t=1}^{T}\sum_i p_i^{(t)}
 =\sum_i c_i^{(T)}
 =\sum_i\max_{t\leq T}b_i^{(t)}
 \leq\sum_i b_i^{\max}.
 \label{eq:bound}
\end{equation}

\begin{table*}[t]
  \centering
  \footnotesize
  \renewcommand{\arraystretch}{1.2}
  \newcommand{\na}{--}
  \newcommand{\hd}[2]{\begin{tabular}{@{}c@{}}#1\\[-1pt]#2\end{tabular}}  %
  \newcommand{\group}[1]{\rowcolor{groupbg}\multicolumn{10}{l}{\textit{#1}} \\}  %
  \newcommand{\maintab}{%
  \begin{tabular}{l cc cc ccccc}
    \toprule
    \multirow{2}{*}{Method} & \multicolumn{2}{c}{MLVU (w/o subtitles)} & \multicolumn{2}{c}{LVB (w/o subtitles)} & \multicolumn{5}{c}{Video-MME (w/ subtitles)} \\
    \cmidrule(lr){2-3} \cmidrule(lr){4-5} \cmidrule(lr){6-10}
    & TFLOPs~$\downarrow$ & Overall~$\uparrow$
      & TFLOPs~$\downarrow$ & Overall~$\uparrow$
      & TFLOPs~$\downarrow$
      & Overall~$\uparrow$ & Short~$\uparrow$ & Medium~$\uparrow$ & Long~$\uparrow$ \\
    \specialrule{\lightrulewidth}{\aboverulesep}{0pt}
    \group{Qwen3-VL-8B~\cite{bai2025qwen3vl}}
    Uniform sampling                                         & 215 & 73.4 & 229 & 65.1 & 224 & 72.9 & \textbf{83.6} & 73.0 & 62.0 \\
    \rowcolor{groupbg}RAG                                                      &  &  &  &  &  &  &  &  &  \\
    \quad w/ Top-K                                  & 218 & 75.9 & 232 & 64.0 & 230 & 73.1 & 82.4 & 73.4 & 63.6 \\
    \quad w/ Q-Frame~\cite{zhang2025qframe}\venue{ICCV'25}    & 209 & 74.7 & 225 & 65.7 & 218 & 71.1 & 80.1 & 70.3 & 63.0 \\
    \quad w/ QuoTA~\cite{luo2025quota}\venue{AAAI'26}         & 781 & 77.2 & 676 & \textbf{66.6} & 647 & 74.4 & 83.2 & 74.4 & 65.4 \\
    \quad w/ FOCUS~\cite{zhu2026focus}\venue{ICLR'26}         & 211 & 71.7 & 226 & 64.0 & 224 & 70.1 & 79.1 & 69.4 & 61.8 \\
    \quad w/ AdaAlloc~\cite{an2026adaalloc}\venue{NeurIPS'26} & 203 & 73.2 & 222 & 65.1 & 225 & 71.0 & 82.8 & 69.4 & 60.7 \\
    \rowcolor{oursbg}\quad w/ \textbf{\name (Ours)}                            & 230 & \textbf{78.1} & 234 & \textbf{66.6} & 232 & \textbf{77.3} & \textbf{83.6} & \textbf{76.8} & \textbf{71.7} \\
    VideoRAG~\cite{ren2025videorag}\venue{KDD'26}             & 211 & 73.8 & 207 & \textbf{61.9} & 220 & 72.3 & \textbf{81.0} & 74.0 & 61.8 \\
    \rowcolor{oursbg}\quad w/ \textbf{\name (Ours)}                            & 208 & \textbf{74.2} & 200 & 61.2 & 217 & \textbf{74.5} & 80.5 & \textbf{74.9} & \textbf{68.0} \\
    \specialrule{\lightrulewidth}{0pt}{0pt}
    \group{InternVL3.5-4B~\cite{wang2025internvl35}}
    Uniform sampling                                         & 143 & 69.9 & 143 & 62.0 & 141 & 65.6 & \textbf{79.8} & 63.8 & 53.2 \\
    \rowcolor{groupbg}RAG                                                      &  &  &  &  &  &  &  &  &  \\
    \quad w/ Top-K                               & 145 & 72.5 & 146 & 60.4 & 145 & 68.6 & 78.0 & 69.1 & 58.7 \\
    \quad w/ Q-Frame~\cite{zhang2025qframe}\venue{ICCV'25}    & 146 & 71.1 & 146 & 63.1 & 142 & 66.0 & 76.7 & 64.6 & 56.9 \\
    \quad w/ QuoTA~\cite{luo2025quota}\venue{AAAI'26}         & 708 & 71.3 & 583 & 62.5 & 551 & 64.9 & 74.7 & 64.6 & 55.4 \\
    \quad w/ FOCUS~\cite{zhu2026focus}\venue{ICLR'26}         & 142 & 69.5 & 143 & 58.8 & 140 & 66.4 & 75.9 & 65.9 & 57.4 \\
    \quad w/ AdaAlloc~\cite{an2026adaalloc}\venue{NeurIPS'26} & 140 & 69.6 & 143 & 61.3 & 139 & 65.7 & 76.3 & 66.4 & 54.2 \\
    \rowcolor{oursbg}\quad w/ \textbf{\name (Ours)}                            & 152 & \textbf{73.0} & 152 & \textbf{63.2} & 150 & \textbf{72.1} & 79.0 & \textbf{70.8} & \textbf{66.4} \\
    \specialrule{\heavyrulewidth}{0pt}{\belowbottomsep}
  \end{tabular}}%
  \sbox0{\setlength{\tabcolsep}{0pt}\maintab}%
  \setlength{\tabcolsep}{\dimexpr(\textwidth-\wd0)/20\relax}%
  \maintab
  \caption{Accuracy at matched compute on three video understanding benchmarks, for all baselines and both backbones.}
  \label{tab:main}
\end{table*}

\paragraph{Attention Refinement.}
Multi-turn conversations offer another unique opportunity to refine relevance using the VLM's attention to previously cached evidence. Since the query must be prefilled in any case, \name prefills $q_t$ before allocating, which gives access to the attention from its query tokens to each cached item at no additional inference cost. Let $a_i^{(t)}$ denote the attention received by item $i$, normalized per cached token. Since attention is unavailable for uncached items, we refine only the relevance of previously seen evidence:
\begin{equation}
\hat r_i^{(t)}=
\begin{cases}
\alpha_t\,r_i(q_t)^{1-\omega}(a_i^{(t)})^\omega,
& i\in\mathcal C_{t-1},\\
r_i(q_t), & \text{otherwise},
\end{cases}
\label{eq:fusion}
\end{equation}
where $\mathcal C_{t-1}$ is the set of cached items, $\omega\in[0,1]$ controls the contribution of attention, and $\alpha_t$ normalizes the fused scores to preserve the total retrieval relevance of cached items. Thus, attention re-ranks previously observed evidence without changing the relevance scores of unseen candidates. Importantly, refinement updates only item relevance, leaving the calibrated token-value functions $v_i(b_i)$ unchanged. The Token Allocator then uses $\hat r_i^{(t)}$ in \cref{eq:allocation}, incrementally prefilling only allocation differences.

\section{Evaluation}
\label{sec:exp}

We evaluate \name on two VLM backbones, Qwen3-VL-8B~\cite{bai2025qwen3vl} and InternVL3.5-4B~\cite{wang2025internvl35}, using Qwen3-VL-8B as the default. As a plug-and-play token allocation framework, \name complements existing video RAG systems rather than replacing them. We evaluate its effectiveness on two retrieval pipelines:
(i) \textit{Embedding-based RAG}, which indexes video frames sampled at 1 FPS alongside temporally aligned subtitles, using SigLIP~2~\cite{siglip2} and BGE~\cite{bge} to score their respective relevance to the query; and
(ii) \textit{Graph-based RAG}, built on VideoRAG~\cite{ren2025videorag}, which retrieves related visual and textual evidence through a graph structure.

\paragraph{Benchmarks.} 

We evaluate on three widely used single-turn, multiple-choice video understanding benchmarks and one multi-turn, open-ended benchmark:
\begin{itemize}\setlength{\itemsep}{0pt}\setlength{\parskip}{0pt}\setlength{\topsep}{2pt}
\item \textbf{MLVU}~\cite{zhou2024mlvu}: 2,174 multiple-choice questions across 1,242 videos from 3 minutes to 2 hours, spanning seven tasks from topic reasoning to action counting.

\item \textbf{LongVideoBench (LVB)}~\cite{wu2024longvideobench}: 1,337 multiple-choice questions across 752 videos from 8 seconds to 1 hour, covering 17 perception and relation tasks.

\item \textbf{Video-MME}~\cite{fu2024videomme}: 2,700 multiple-choice questions across 900 videos, divided into short ($<$2 min), medium (4--15 min), and long (30--60 min) subsets.

\item \textbf{MT-Video-Bench (MTVB)}~\cite{mtvideobench2025}: 1,000 open-ended dialogues (5--8 turns) across 135 videos from 20 seconds to 41 minutes. Open-ended responses are scored using an LLM judge (Gemini 3.8 Flash) against a provided answer checklist.
\end{itemize}

\paragraph{Baselines.}
We compare \name against uniform sampling and five recent frame selection and token allocation methods, all operating under the same token budget. 
\begin{itemize}\setlength{\itemsep}{0pt}\setlength{\parskip}{0pt}\setlength{\topsep}{2pt}
\item \textbf{Top-K}: Retrieves the $K$ frames with the highest query--frame semantic similarity scores.

\item \textbf{FOCUS}~\cite{zhu2026focus}: Selects $K$ frames through multi-armed bandit exploration over temporal clips.

\item \textbf{Q-Frame}~\cite{zhang2025qframe}: Samples frames according to query relevance and allocates tokens across high-, medium-, and low-resolution tiers with a 2:1:1 budget ratio.

\item \textbf{QuoTA}~\cite{luo2025quota}: Rewrites queries into relevant entities and uses a VLM to score each frame by its likelihood of containing those entities. It then allocates tokens proportionally to the scores and downsamples tokens of each frame to a particular resolution.

\item \textbf{AdaAlloc}~\cite{an2026adaalloc}: Uses an agent to plan how to distribute the budget between low- and high-resolution frame tiers.
\end{itemize}
For multi-turn evaluation, we enable KV-cache reuse whenever a method selects evidence that exactly matches previously prefilled content: Top-K and FOCUS reuse previously seen frames, while Q-Frame and AdaAlloc reuse frames only at matching resolutions. QuoTA does not support this reuse because it downsamples tokens to variable amounts, which makes reuse impossible.

\paragraph{Metrics.}
We report answering accuracy and inference compute (TFLOPs), with the latter controlled through the token budget. To ensure a fair comparison between single- and multi-call methods, we account for prefill compute across all model calls, including auxiliary frame scoring and planning calls. Offline indexing, including frame encoding and token ranking, runs once per video and is shared by all queries, so we exclude it for all methods. Experiments are conducted on A100 GPUs. Additional implementation details, hyperparameters, and sensitivity analyses are provided in the \suppmaterial.

\subsection{Main Results}
\label{sec:exp-main}

\begin{figure}[t]
  \centering
  \includegraphics[width=\linewidth]{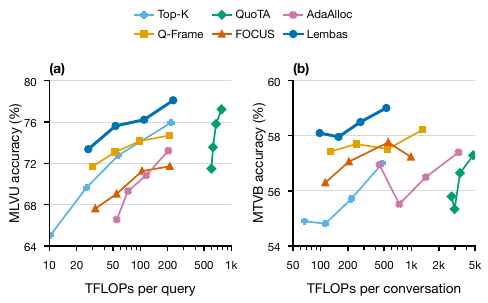}
  \caption{\name achieves the compute-to-accuracy Pareto frontier across (a) single-turn and (b) multi-turn video understanding.
  }
  \label{fig:multiturn}
\end{figure}

\paragraph{\name achieves the Pareto frontier across single-turn video understanding.} 
As shown in \cref{tab:main}, \name achieves the highest overall accuracy across all three benchmarks and both VLM backbones. With Qwen3-VL-8B, it improves over Top-K by 2.2, 2.6, and 4.2 percentage points on MLVU, LVB, and Video-MME, respectively, at comparable compute. The gains are particularly pronounced on long videos: on the long subset of Video-MME, \name improves accuracy from 63.6\% to 71.7\% with nearly identical TFLOPs. QuoTA achieves competitive accuracy but incurs 2.8--3.4$\times$ higher compute due to its VLM-based frame scoring. These trends generalize to InternVL3.5-4B, where \name improves over Top-K by 0.5--3.5 points. Furthermore, \name consistently achieves favorable accuracy--compute trade-offs across multiple token budgets (\cref{fig:multiturn}a).

When integrated with VideoRAG, \name improves accuracy by 0.4 and 2.2 percentage points on MLVU and Video-MME, respectively, by allocating tokens across more frames at comparable compute, while accuracy decreases by 0.7 points on LVB. These results demonstrate that \name complements existing RAG designs.

\paragraph{\name achieves the Pareto frontier across multi-turn video understanding.} 
As shown in \cref{fig:multiturn}b, \name outperforms all baselines by 0.3--1.5 percentage points while requiring 1.7--17.3$\times$ fewer TFLOPs. Unlike prior methods that reuse only previously selected frames or exact resolution matches, \name incrementally prefills only tokens not already present in the KV cache. This finer-grained reuse substantially reduces redundant computation across conversational turns while preserving answer accuracy.

\paragraph{\name introduces negligible compute overhead.}
Semantic similarity search with FAISS~\cite{faiss}, iterative water-filling, and token pruning together run on the order of tens of milliseconds per query, which is negligible compared with VLM inference.

\begin{table}[t]
  \centering
  \footnotesize
  \renewcommand{\arraystretch}{1.2}
  \settowidth{\splitw}{Medium}%
  \newcommand{\na}{--}
  \newcommand{\abltab}{%
  \begin{tabular}{l<{\hspace{5pt}} c VVVV}
    \toprule
    Method & TFLOPs $\downarrow$
      & \multicolumn{4}{c}{%
        \begin{tabular}{@{}VVVV@{}}
          \multicolumn{4}{c}{Video-MME (w/ subtitles) $\uparrow$} \\
          \cmidrule(lr){1-4}
          Overall & Short & Medium & Long \\[2.6pt]
        \end{tabular}} \\
    \specialrule{\lightrulewidth}{\aboverulesep}{0pt}
    \rowcolor{oursbg}\textbf{\name}                 & 232 & \textbf{77.3} & \textbf{83.6} & \textbf{76.8} & \textbf{71.7} \\
    \hspace{0.5em}w/o text allocation      & 226 & 73.7 & 81.9 & 74.9 & 64.4 \\
    \hspace{0.5em}w/o diversity term            & 226 & 76.6 & 82.4 & 76.4 & 70.8 \\
    \hspace{0.5em}w/o water-filling        & 229 & 76.1 & 82.3 & 76.1 & 69.9 \\
    \bottomrule
  \end{tabular}}%
  \sbox0{\setlength{\tabcolsep}{0pt}\abltab}%
  \setlength{\tabcolsep}{\dimexpr(\columnwidth-\wd0)/11\relax}%
  \xdef\ablsep{\the\tabcolsep}%
  \abltab
  \caption{Ablation studies across single-turn Q\&A.}
  \label{tab:ablation}

  \vspace{8pt}  %

  \centering
  \footnotesize
  \renewcommand{\arraystretch}{1.2}
  \setlength{\tabcolsep}{\ablsep}%
  \settowidth{\dimen0}{\hspace{0.5em}w/o incremental allocation}%
  \dimen2=\dimexpr(\columnwidth-5\tabcolsep-5pt-\dimen0)/2\relax
  \begin{tabular}{l<{\hspace{5pt}} >{\centering\arraybackslash}p{\dimen2} >{\centering\arraybackslash}p{\dimen2}}
    \toprule
    Method & TFLOPs $\downarrow$ & MTVB $\uparrow$ \\
    \specialrule{\lightrulewidth}{\aboverulesep}{0pt}
    \rowcolor{oursbg}\textbf{\name}             & 275 & 58.5 \\
    \hspace{0.5em}w/o attention refinement       & 315 & 57.4 \\
    \hspace{0.5em}w/o incremental allocation & 687 & \textbf{59.1} \\
    \bottomrule
  \end{tabular}
  \caption{Ablation studies across multi-turn Q\&A.}
  \label{tab:multiturn}
\end{table}

\paragraph{Performance breakdown.}
We ablate the key components of \name in \cref{tab:ablation,tab:multiturn}. On Video-MME, removing subtitle allocation decreases overall accuracy by 3.6 points, with a particularly large 7.3-point drop on long videos. Even without subtitle allocation, \name outperforms all similarly budgeted, non-agentic frame selection baselines by 0.6--3.6 points. Removing the diversity term reduces accuracy by 0.7 points, confirming the benefit of modeling redundancy, and replacing water-filling with a simpler proportional allocation, which offers no optimality guarantee, reduces it by 1.2 points.

For multi-turn conversations, removing attention refinement decreases accuracy by 1.1 points while increasing prefill compute from 275 to 315 TFLOPs. Disabling incremental allocation increases compute from 275 to 687 TFLOPs (2.5$\times$), with only a 0.6-point accuracy increase. These results show that attention refinement improves evidence selection, while incremental allocation is the primary driver of multi-turn compute savings. 

\section{Conclusion}
\label{sec:conclusion}

We introduce \name, a plug-and-play framework for token-efficient video RAG that jointly allocates a shared token budget across frames and subtitles. Its Token Allocator optimizes query relevance, diminishing token value, and evidence diversity through iterative water-filling, while incremental allocation and attention refinement enable efficient reuse across conversational turns. At comparable compute, \name improves accuracy by 0.1--7.2 percentage points over state-of-the-art baselines. On multi-turn video understanding, \name achieves 0.3--1.5 percentage points higher accuracy with 1.7--17.3$\times$ fewer TFLOPs.

\section*{LLM Usage Statement}
We were assisted by various LLMs, specifically from the ChatGPT, Claude, and Gemini model families in preparing this work. They were used to assist in paper writing, related work search, and coding which the authors reviewed and edited. LLMs were used to a lesser extent in the design and analysis of core ideas and experiments. Gemini 3.8 Flash was used to evaluate multi-turn benchmark results. LLMs were not used to create or annotate data. The authors take full responsibility for all content.

{
    \small
    \bibliographystyle{ieeenat_fullname}
    \bibliography{main}
}

\clearpage
\providecommand{\mainpaper}{main paper}
\ifdefined\appendixmode
  \appendix
  \section*{Appendix}
  This appendix provides the derivations and proofs omitted from the \mainpaper{} (\cref{sec:supp-proofs}), implementation details (\cref{sec:supp-impl}), and additional results (\cref{sec:supp-results}).
\else
  \maketitlesupplementary
  \setcounter{figure}{8}\setcounter{table}{3}\setcounter{equation}{11}
  \setcounter{section}{0}\renewcommand{\thesection}{\Alph{section}}

  In this supplementary material, we provide the derivations and proofs omitted from the \mainpaper{} (\cref{sec:supp-proofs}), implementation details (\cref{sec:supp-impl}), and additional results (\cref{sec:supp-results}).
\fi
\raggedbottom  %
\renewcommand{\qedsymbol}{}\renewcommand{\qed}{}  %

\section{Derivations and Proofs}
\label{sec:supp-proofs}

For completeness, we first restate the allocation problem of the \mainpaper. Let $\mathcal F$ and $\mathcal S$ be the candidate frames and subtitles, let $b_i\in[0,b_i^{\max}]$ be the token count of item $i\in\mathcal F\cup\mathcal S$, and let $v_i(b_i)=b_i/(b_i+\tau_i)$ be its token value, where $\tau_i$ is $\tau_{\text{frame}}$ for a frame and $\tau_{\text{text}}$ for a subtitle. \name solves
\begin{equation}
\begin{aligned}
  \max_{\mathbf b}\quad & \sum_{i\in\mathcal F} r_i(q,\mathbf b)\,v_i(b_i)+\beta\sum_{i\in\mathcal S} r_i(q,\mathbf b)\,v_i(b_i)\\
  \text{s.t.}\quad & \sum_{i\in\mathcal F\cup\mathcal S} b_i\le B ,
\end{aligned}
\label{eq:supp-allocation}
\end{equation}
where the allocation-dependent relevance $r_i(q,\mathbf b)$ discounts the retrieval relevance $r_i(q)$ of item $i$ by the token values of similar items in its modality $\mathcal M_i\in\{\mathcal F,\mathcal S\}$:
\begin{equation}
  r_i(q,\mathbf b)=r_i(q)-\eta\sum_{j\in\mathcal M_i}s_{ij}\,v_j(b_j) .
\label{eq:supp-cond}
\end{equation}
Here, $s_{ij}$ is the cosine similarity between the embeddings of items $i$ and $j$, and $\eta\ge0$ is the redundancy penalty.

\subsection{NP-Hardness of the Discrete Problem}
\label{sec:supp-nphard}

The VLM consumes whole tokens, so in practice every token count $b_i$ must be an integer in $\{0,1,\dots,b_i^{\max}\}$. We show that this discrete version of \cref{eq:supp-allocation} is NP-hard. Its decision version asks whether some integer allocation within the budget $B$ reaches an objective value of at least $\theta$.

\smallskip
\noindent\textbf{Proposition 2.} \textit{Maximizing \cref{eq:supp-allocation} over integer token counts is NP-hard, even when every item can receive at most one token.}

\smallskip
We reduce from the independent set problem, which is NP-complete~\cite{karp1972,garey1979}: given a graph $G=(V,E)$ with $n$ vertices and an integer $k$, decide whether $G$ contains $k$ vertices, no two of which are adjacent. The idea is that the redundancy penalty acts as a penalty on the edges of $G$. Each vertex becomes a frame, and two adjacent frames that both receive a token lower the objective, so the best $k$ frames form an independent set whenever one exists.

\begin{proof}
\textit{Construction.} Given $G$ and $k$, we build an instance with one frame per vertex and no subtitles:
\begin{equation}
\begin{aligned}
  &b_i^{\max}=1,\quad \tau_i=1,\quad r_i(q)=\tfrac1n,\\
  &\eta=\tfrac1n,\quad B=k,\quad \mathbf S=\mathbf I+\tfrac1n\mathbf A ,
\end{aligned}
\label{eq:supp-reduction}
\end{equation}
where $\mathbf A$ is the adjacency matrix of $G$. Two adjacent frames thus have similarity $1/n$, and all other pairs of distinct frames have similarity $0$. The construction takes polynomial time.

\textit{Validity.} The objective depends on the embeddings only through their cosine similarities, so it suffices that $\mathbf S$ is the matrix of cosine similarities of some unit vectors. Every eigenvalue of $\mathbf A$ has magnitude at most the maximum degree of $G$, which is at most $n-1$, so every eigenvalue of $\mathbf S$ is at least $1/n>0$. Therefore, $\mathbf S$ is positive definite with unit diagonal, i.e., it is the Gram matrix of $n$ unit vectors.

\textit{Objective.} Each frame receives either no token, with token value $0$, or one token, with token value $v_i(1)=1/2$. Let $X$ be the set of frames that receive a token, and let $e(X)$ be the number of edges of $G$ with both endpoints in $X$. The budget requires $|X|\le k$, and the objective of \cref{eq:supp-allocation} becomes
\begin{equation}
\begin{aligned}
  J(X)&=\frac{|X|}{2n}-\frac{1}{4n}\sum_{i,j\in X}s_{ij}\\
      &=\frac{|X|}{2n}-\frac{|X|}{4n}-\frac{e(X)}{2n^2}
       =\frac{|X|}{4n}-\frac{e(X)}{2n^2} .
\end{aligned}
\label{eq:supp-nphard}
\end{equation}

\textit{Correctness.} If $G$ contains an independent set of size $k$, giving one token to each of its frames is feasible and attains $J=k/(4n)$. Conversely, every feasible allocation satisfies $J(X)\le|X|/(4n)\le k/(4n)$, with equality only if $|X|=k$ and $e(X)=0$, i.e., only if $X$ is an independent set of size $k$. Hence, $G$ contains an independent set of size $k$ if and only if the discrete problem reaches $\theta=k/(4n)$. A polynomial-time algorithm for the discrete problem would thus decide the independent set problem in polynomial time.
\end{proof}

\noindent The construction relies on the redundancy penalty, which couples the items: without it ($\eta=0$), giving one token to any $k$ frames is optimal. In the continuous relaxation, the same coupling still yields a convex program, because $\mathbf S$ is positive semidefinite (\cref{sec:supp-convex}). \name therefore solves the relaxation and rounds its solution down to whole tokens.

\subsection{Derivation of Water-Filling}
\label{sec:supp-waterfill}

We derive the water-filling allocation of the \mainpaper{} as the unique solution of the subproblem in which the relevance of each item is fixed to $\bar r_i>0$:
\begin{equation}
\begin{aligned}
  \max_{\mathbf b}\quad & \sum_i \bar r_i\,\frac{b_i}{b_i+\tau_i}\\
  \text{s.t.}\quad & \sum_i b_i\le B,\quad 0\le b_i\le b_i^{\max} .
\end{aligned}
\label{eq:supp-fixed}
\end{equation}
We solve this subproblem through its Lagrangian dual. The idea is to replace the budget with a price per token, let each item acquire tokens at that price, and then adjust the price until the budget is spent. We introduce a multiplier $\lambda\ge0$ for the budget constraint and keep the bounds $0\le b_i\le b_i^{\max}$ as the domain of $\mathbf b$, which gives the Lagrangian
\begin{equation}
  \mathcal L(\mathbf b,\lambda)=\sum_i \bar r_i\,\frac{b_i}{b_i+\tau_i}-\lambda\Big(\sum_i b_i-B\Big) .
\label{eq:supp-lagrangian}
\end{equation}
The subproblem is equivalent to $\max_{\mathbf b}\min_{\lambda\ge0}\mathcal L(\mathbf b,\lambda)$, and its dual is $\min_{\lambda\ge0}\max_{\mathbf b}\mathcal L(\mathbf b,\lambda)$. Strong duality holds because the objective is concave, the constraints are linear, and $\mathbf b=\mathbf 0$ is strictly feasible~\cite{boyd2004}. Hence, the two problems attain the same value, and the solution of the subproblem maximizes $\mathcal L(\cdot,\lambda^\star)$ at the dual optimum $\lambda^\star$.

\paragraph{Inner maximization.}
For a fixed $\lambda>0$, the Lagrangian separates over the items, and each item solves
\begin{equation}
  \max_{0\le b_i\le b_i^{\max}}\ \bar r_i\,\frac{b_i}{b_i+\tau_i}-\lambda\,b_i .
\label{eq:supp-item}
\end{equation}
This objective is strictly concave. Setting its derivative to zero gives
\begin{equation}
  \frac{\bar r_i\,\tau_i}{(b_i+\tau_i)^2}=\lambda ,
\label{eq:supp-stationarity}
\end{equation}
and clipping the solution to the domain gives the unique maximizer
\begin{equation}
  b_i(\lambda)=\left[\sqrt{\bar r_i\tau_i/\lambda}-\tau_i\right]_0^{b_i^{\max}} ,
\label{eq:supp-waterfill}
\end{equation}
which is the water-filling allocation of the \mainpaper. Intuitively, each item acquires tokens until its marginal gain falls to the price $\lambda$. An item thus receives no tokens when its marginal gain at zero tokens, $\bar r_i/\tau_i$, does not exceed $\lambda$, and it receives its full token count $b_i^{\max}$ when its marginal gain there is still at least $\lambda$. In iterative water-filling, the discounted relevance of an item can be non-positive. Such an item cannot increase the objective and receives no tokens.

\paragraph{Outer minimization.}
Substituting \cref{eq:supp-waterfill} into \cref{eq:supp-lagrangian} gives the dual function $g(\lambda)=\mathcal L(\mathbf b(\lambda),\lambda)$, which is convex in the single variable $\lambda$ with derivative
\begin{equation}
  g'(\lambda)=B-\sum_i b_i(\lambda) .
\label{eq:supp-dual}
\end{equation}
If $\sum_i b_i^{\max}\le B$, the budget is not binding and every item receives its full token count. Otherwise, the dual optimum $\lambda^\star$ satisfies $g'(\lambda^\star)=0$, i.e., the allocation spends exactly the budget. Because of the clipping, this equation has no closed-form solution in general. However, each $b_i(\lambda)$ is continuous and non-increasing in $\lambda$, and all of them vanish once $\lambda\ge\max_i\bar r_i/\tau_i$. A bisection over $\lambda\in(0,\max_i\bar r_i/\tau_i]$ therefore finds $\lambda^\star$. Since the inner maximizer is unique, the resulting allocation $\mathbf b(\lambda^\star)$ is the unique solution of \cref{eq:supp-fixed}.

\subsection{Convergence of Iterative Water-Filling}
\label{sec:supp-convergence}
\label{sec:supp-convex}

We now prove Proposition~1 of the \mainpaper, which bounds the gap between iterative water-filling and the global maximum of \cref{eq:supp-allocation}. We first show that the continuous relaxation is a convex program and then prove the bound.

\paragraph{Convexity of the continuous relaxation.}
The continuous relaxation lets each token count $b_i$ take any real value in $[0,b_i^{\max}]$. After a change of variables, this relaxation maximizes a concave function over a convex set.

We change variables from the token counts to the token values $z_i=v_i(b_i)$, and write $\mathbf b$ and $\mathbf z$ for the vectors over all items. Since $v_i$ is strictly increasing, this change of variables is invertible, with
\begin{equation}
  b_i=v_i^{-1}(z_i)=\frac{\tau_i z_i}{1-z_i} ,
\label{eq:supp-inverse}
\end{equation}
where $0\le z_i\le z_i^{\max}=v_i(b_i^{\max})<1$.
In these variables, the budget $B$ restricts the token values to the feasible set
\begin{equation}
  \mathcal Z=\Big\{\mathbf z:\ 0\le z_i\le z_i^{\max},\ \ \sum_{i}\frac{\tau_i z_i}{1-z_i}\le B\Big\} .
\label{eq:supp-feasible}
\end{equation}
Each function $\tau_i z_i/(1-z_i)$ is convex on $[0,1)$, because its second derivative $2\tau_i/(1-z_i)^3$ is positive. The budget constraint thus keeps a convex function below $B$, so $\mathcal Z$ is convex, and it is compact because it is closed and bounded.

The objective, in turn, is concave in the token values. Substituting \cref{eq:supp-cond} into \cref{eq:supp-allocation} gives
\begin{equation}
  J(\mathbf z)=\bar{\mathbf r}^{\top}\mathbf z-\eta\,\mathbf z^{\top}\mathbf S\mathbf z ,
\label{eq:supp-quadratic}
\end{equation}
where $\bar r_i=r_i(q)$ for a frame and $\bar r_i=\beta\,r_i(q)$ for a subtitle, and the similarity matrix $\mathbf S$ is block-diagonal, with the frame similarity matrix and the $\beta$-scaled subtitle similarity matrix as its blocks. Each block contains the cosine similarities between the embeddings of one modality, including the diagonal entries $s_{ii}=1$. The frame block is therefore a Gram matrix of normalized embeddings and the subtitle block is a positive multiple of such a matrix, so $\mathbf S$ is positive semidefinite. Since $\eta\ge0$, $J$ is concave. Maximizing $J$ over $\mathcal Z$ is thus a convex program, in which every local maximum is a global maximum~\cite{boyd2004}.

\smallskip
\noindent\textbf{Proposition 1} (restated)\textbf{.} \textit{Let $J^\star$ be the global maximum of \cref{eq:supp-allocation} over continuous token allocations, and let $J^{(\ell)}$ be the objective value after $\ell$ iterations of Algorithm~1 of the \mainpaper. Then}
\begin{equation}
  J^\star-J^{(\ell)}\le\frac{4\,\eta\,\sigma_{\max}D^2}{\ell+2}=O\Big(\frac{1}{\ell}\Big) ,
\label{eq:supp-rate}
\end{equation}
\textit{where $\sigma_{\max}$ is the largest eigenvalue of the frame and $\beta$-scaled subtitle similarity matrices and $D$ is the diameter of the feasible token-value set under budget $B$.}

\smallskip
The proof follows the standard analysis of the Frank--Wolfe method~\cite{jaggi2013}, applied to the token values $\mathbf z$. It uses the gradient $\nabla J(\mathbf z)=\bar{\mathbf r}-2\eta\,\mathbf S\mathbf z$, which is the discounted relevance $\tilde{\mathbf r}$ of Eq.~(6) of the \mainpaper, and the constant $C=2\eta\,\sigma_{\max}$, which bounds the curvature of $J$. The diameter $D$ is the largest distance between two points of $\mathcal Z$. The proof rests on two lemmas. The first bounds how far the current token values are from the optimum, and the second bounds how much one iteration improves the objective.

\begin{lemma}[Distance to the optimum]
\label{lem:gap}
For every $\mathbf z\in\mathcal Z$,
\begin{equation}
  J^\star-J(\mathbf z)\le\max_{\mathbf y\in\mathcal Z}\,\nabla J(\mathbf z)^{\top}(\mathbf y-\mathbf z) .
\label{eq:supp-lem-gap}
\end{equation}
\end{lemma}
\begin{proof}
Let $\mathbf z^{\rm opt}$ be a maximizer of $J$ over $\mathcal Z$. Since $J$ is concave, its linear approximation at $\mathbf z$ lies above it, so $J^\star=J(\mathbf z^{\rm opt})\le J(\mathbf z)+\nabla J(\mathbf z)^{\top}(\mathbf z^{\rm opt}-\mathbf z)$. The right-hand side is at most the maximum over all $\mathbf y\in\mathcal Z$.
\end{proof}

\noindent Intuitively, the best improvement that the linear approximation predicts is never smaller than the remaining gap. One water-filling pass solves exactly this maximization: in token values, the objective of \cref{eq:supp-fixed} with relevances $\nabla J(\mathbf z)$ is the linear function $\nabla J(\mathbf z)^{\top}\mathbf y$, and its constraints define $\mathcal Z$.

\begin{lemma}[Progress of one step]
\label{lem:step}
For every $\mathbf z,\mathbf s\in\mathcal Z$ and $\gamma\in[0,1]$,
\begin{equation}
\begin{aligned}
  J\big(\mathbf z+\gamma(\mathbf s-\mathbf z)\big)\ge{}& J(\mathbf z)+\gamma\,\nabla J(\mathbf z)^{\top}(\mathbf s-\mathbf z)\\
  &-\frac{C}{2}\gamma^2D^2 .
\end{aligned}
\label{eq:supp-lem-step}
\end{equation}
\end{lemma}
\begin{proof}
Write $\mathbf d=\gamma(\mathbf s-\mathbf z)$. Since $J$ is quadratic, its second-order expansion is exact: $J(\mathbf z+\mathbf d)=J(\mathbf z)+\nabla J(\mathbf z)^{\top}\mathbf d-\eta\,\mathbf d^{\top}\mathbf S\mathbf d$. The last term is at least $-\eta\,\sigma_{\max}\|\mathbf d\|^2=-\frac{C}{2}\|\mathbf d\|^2$, and $\|\mathbf d\|\le\gamma D$.
\end{proof}

\noindent That is, a step of size $\gamma$ toward $\mathbf s$ gains at least a fraction $\gamma$ of the improvement predicted for a full step, minus a curvature term that shrinks with $\gamma^2$.

\begin{proof}[Proof of Proposition 1]
We first derive a recursion for the remaining gap and then solve it by induction. Let $\mathbf z^{(\ell-1)}$ be the token values before iteration $\ell$, with $\mathbf z^{(0)}=\mathbf 0$, and let $h_\ell=J^\star-J(\mathbf z^{(\ell)})$ be the remaining gap. Iteration $\ell$ runs water-filling with the discounted relevance $\nabla J(\mathbf z^{(\ell-1)})$, which returns the maximizer $\mathbf s$ of the right-hand side of \cref{eq:supp-lem-gap} ($\mathbf z^\star$ in the \mainpaper), and moves to $\mathbf z^{(\ell)}=\mathbf z^{(\ell-1)}+\gamma_\ell(\mathbf s-\mathbf z^{(\ell-1)})$ with $\gamma_\ell=2/(\ell+1)$. This point lies in $\mathcal Z$ because $\mathcal Z$ is convex.

Applying \cref{lem:step} with $\mathbf z=\mathbf z^{(\ell-1)}$ and $\gamma=\gamma_\ell$ gives
\begin{align*}
  J(\mathbf z^{(\ell)})
  &\ge J(\mathbf z^{(\ell-1)})+\gamma_\ell\,\nabla J(\mathbf z^{(\ell-1)})^{\top}(\mathbf s-\mathbf z^{(\ell-1)})\\
  &\qquad-\frac{C}{2}\gamma_\ell^2D^2\\
  &\ge J(\mathbf z^{(\ell-1)})+\gamma_\ell\,h_{\ell-1}-\frac{C}{2}\gamma_\ell^2D^2 ,
\end{align*}
where the second inequality uses \cref{lem:gap}. Subtracting both sides from $J^\star$ gives
\begin{equation}
  h_\ell\le(1-\gamma_\ell)\,h_{\ell-1}+\frac{C}{2}\gamma_\ell^2D^2 .
\label{eq:supp-recursion}
\end{equation}
Each iteration thus removes a fraction $\gamma_\ell$ of the remaining gap and adds a curvature term of order $\gamma_\ell^2$.

We now show $h_\ell\le2CD^2/(\ell+2)$ by induction. For $\ell=1$, the step size is $\gamma_1=1$, and \cref{eq:supp-recursion} gives $h_1\le CD^2/2\le2CD^2/3$. For $\ell\ge2$, assume the claim for $\ell-1$, i.e., $h_{\ell-1}\le2CD^2/(\ell+1)$. Then
\begin{align*}
  h_\ell&\overset{\text{(a)}}{\le}\frac{\ell-1}{\ell+1}\,h_{\ell-1}+\frac{2CD^2}{(\ell+1)^2}\\
  &\overset{\text{(b)}}{\le}\frac{\ell-1}{\ell+1}\cdot\frac{2CD^2}{\ell+1}+\frac{2CD^2}{(\ell+1)^2}\\
  &=\frac{2CD^2\,\ell}{(\ell+1)^2}\overset{\text{(c)}}{\le}\frac{2CD^2}{\ell+2} ,
\end{align*}
where (a) substitutes $\gamma_\ell=2/(\ell+1)$ into \cref{eq:supp-recursion}, (b) applies the induction hypothesis, and (c) uses $\ell(\ell+2)\le(\ell+1)^2$. Substituting $C=2\eta\,\sigma_{\max}$ gives \cref{eq:supp-rate}, which completes the proof.
\end{proof}

\noindent Proposition~1 has two implications. First, without the redundancy penalty ($\eta=0$), the bound is zero, so the first iteration, which is plain water-filling, is already optimal. Second, since every token value lies in $[0,1)$, the diameter satisfies $D^2\le|\mathcal F|+|\mathcal S|$, so the numerator of the bound is at most $4\,\eta\,\sigma_{\max}(|\mathcal F|+|\mathcal S|)$. The bound holds for the continuous allocation, which \name rounds down to whole tokens afterward.

\subsection{Suboptimality of Prior Allocation Rules}
\label{sec:supp-prior}

The \mainpaper{} states that neither proportional allocation~\cite{luo2025quota} nor tiered allocation~\cite{zhang2025qframe} maximizes the fixed-relevance objective of \cref{eq:supp-fixed}. This section proves both statements.

\begin{lemma}[Proportional allocation]
\label{lem:proportional}
Suppose at least three items have pairwise distinct relevances $\bar r_i>0$ and share the same parameter $\tau$, and let $b_i=c\,\bar r_i$ with $c>0$ satisfy $0<b_i<b_i^{\max}$ for these items. Then this allocation does not maximize \cref{eq:supp-fixed}.
\end{lemma}
\begin{proof}
By \cref{sec:supp-waterfill}, an optimal allocation requires all items with $0<b_i<b_i^{\max}$ to share the same marginal gain, i.e., $\bar r_i\tau/(c\,\bar r_i+\tau)^2=\lambda$ for some $\lambda>0$. Rearranging this equation shows that every such relevance is a root of the quadratic $\lambda c^2x^2+(2\lambda c\tau-\tau)x+\lambda\tau^2=0$ in $x$. A quadratic has at most two distinct roots, which contradicts the three pairwise distinct relevances.
\end{proof}

\noindent Intuitively, proportional allocation would equalize the marginal gains only if the token value grew like $\log b_i$, which never saturates. With a saturating token value, proportional allocation keeps adding tokens to the most relevant items after their token value has saturated, and takes these tokens from less relevant items that would still benefit.

\begin{lemma}[Tiered allocation]
\label{lem:tiered}
Suppose two items of the same tier have distinct relevances $\bar r_i\ne\bar r_j$, share the same parameter $\tau$, and receive the same token count $b$ with $0<b<\min(b_i^{\max},b_j^{\max})$. Then this allocation does not maximize \cref{eq:supp-fixed}.
\end{lemma}
\begin{proof}
An optimal allocation requires equal marginal gains for the two items, $\bar r_i\tau/(b+\tau)^2=\bar r_j\tau/(b+\tau)^2$, which holds only if $\bar r_i=\bar r_j$.
\end{proof}

\noindent Intuitively, the more relevant of the two items still gains more from its next token, so moving one token to it raises the objective.

In both cases, the water-filling allocation of \cref{eq:supp-waterfill} is the unique maximizer and therefore attains a strictly higher objective within the same budget.

\section{Implementation Details}
\label{sec:supp-impl}

\subsection{Settings of \name}
\label{sec:supp-settings}

\Cref{tab:supp-hparams} lists the settings of \name. SigLIP~2 embeds the query and the candidate frames, and BGE embeds the query and the candidate subtitles. Within each modality, we standardize the cosine similarities between the query and the candidate items to zero mean and unit variance and convert them into the retrieval relevance $r_i(q)$ with a softmax at temperature $T$. The token value parameters $\tau_{\text{frame}}$ and $\tau_{\text{text}}$ set how quickly the token value of an item saturates and are fitted offline by least squares to the validation measurements of \S3.2 of the \mainpaper. The subtitle weight $\beta$ scales the relevance of subtitles relative to that of frames, and the redundancy penalty $\eta$ scales the discount that an item receives from similar items. Iterative water-filling runs for $L$ iterations and finds $\lambda$ in each iteration by bisection. In multi-turn conversations, the attention weight $\omega$ controls the contribution of attention to the relevance of cached items. The token budget $B$ sets the compute of each experiment, which we report in TFLOPs with each result.

\begin{table}[!htb]
  \centering
  \footnotesize
  \renewcommand{\arraystretch}{1.15}
  \setlength{\tabcolsep}{4pt}
  \begin{tabular}{l l}
    \toprule
    Setting & Value \\
    \midrule
    Frame relevance encoder & SigLIP 2 (so400m-patch14-384)~\cite{siglip2} \\
    Subtitle relevance encoder & BGE (base-en-v1.5)~\cite{bge} \\
    Relevance temperature $T$ & 0.63 \\
    \addlinespace[3pt]
    Token value parameter $\tau_{\text{frame}}$ & 38 \\
    Token value parameter $\tau_{\text{text}}$ & 13.6 \\
    Subtitle weight $\beta$ & 0.5 \\
    Redundancy penalty $\eta$ & $2.67\times10^{-5}$ \\
    \addlinespace[3pt]
    Iterations $L$ & 20 \\
    Bisection steps per iteration & 60 \\
    \addlinespace[3pt]
    Attention weight $\omega$ (multi-turn) & 0.5 \\
    \bottomrule
  \end{tabular}
  \caption{Settings of \name.}
  \label{tab:supp-hparams}
\end{table}

\subsection{Answer Prompt}
\label{sec:supp-prompt}

All methods answer the multiple-choice questions with the prompt in \cref{fig:supp-prompt}. Each frame is preceded by its timestamp, and on Video-MME the selected subtitles precede the question. The prediction is the option letter to which the VLM assigns the highest probability.

\begin{figure}[!htb]
\begin{tcolorbox}[colframe=gray, colback=white, colbacktitle=gray!10, coltitle=black,
    boxrule=0.4mm, arc=1mm, left=4pt, right=4pt, top=3pt, bottom=3pt,
    fonttitle=\bfseries\small, title={Prompt: Multiple-choice answering}]
\small
\textcolor{gray}{\textit{$\langle$4.0 seconds$\rangle$ [frame] \ $\langle$12.0 seconds$\rangle$ [frame] \ \dots}}

\smallskip
Subtitles of the video: \textcolor{gray}{\textit{[subtitles]}}

\smallskip
Select the best answer to the following multiple-choice question based on the video. Respond with only the letter of the correct option.

\smallskip
Question: \textcolor{gray}{\textit{[question]}}\\
Options:\\
A. \textcolor{gray}{\textit{[option]}}\\
B. \textcolor{gray}{\textit{[option]}}\\
\textcolor{gray}{\dots}

\smallskip
The best answer is:
\end{tcolorbox}
\vspace{-2mm}
\caption{Answer prompt shared by all methods. Gray italic text marks the variable inputs of each question.}
\label{fig:supp-prompt}
\end{figure}

\begin{figure*}[t]
  \centering
  \includegraphics[width=\linewidth]{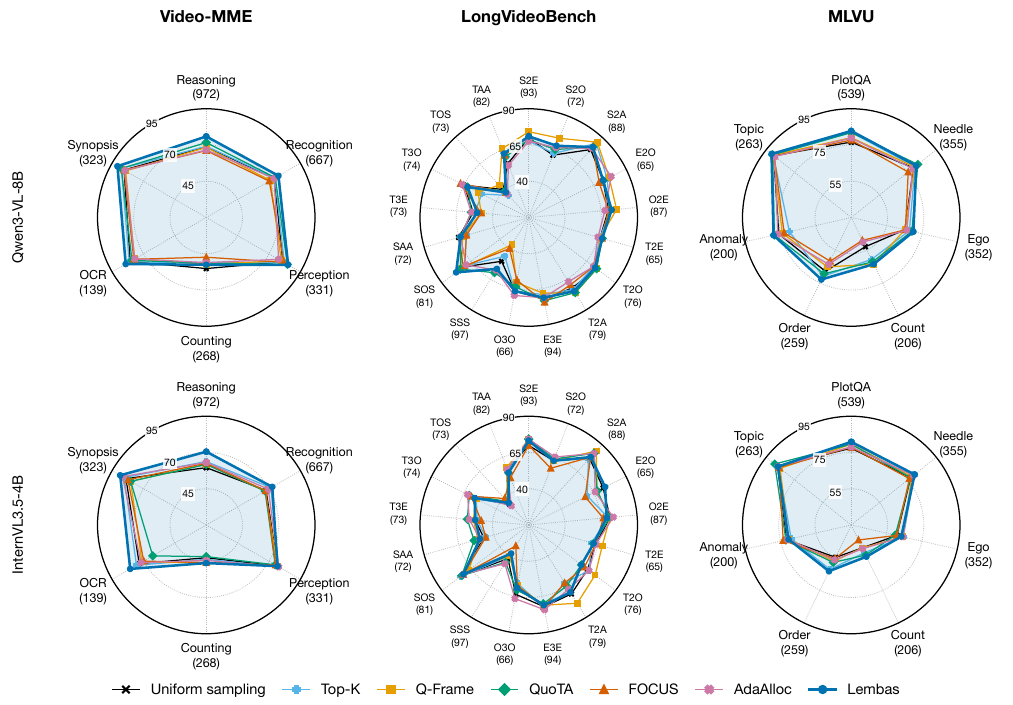}
  \caption{\textbf{\name achieves the highest accuracy in more question categories than any baseline.} Accuracy (\%) per question category, with question counts in parentheses.}
  \label{fig:supp-radar}
\end{figure*}

\subsection{Baseline Implementations}
\label{sec:supp-baselines}

We compare all methods under the same conditions, so that differences in accuracy come from which evidence each method selects and how it divides the token budget. All methods select from the same candidate frames and answer with the same VLM and prompt (\cref{sec:supp-prompt}) within the same token budget. On Video-MME, every baseline also receives the subtitles aligned with the frames it selects, counted within that budget. Every method that scores query--frame relevance with an image--text encoder uses the same SigLIP~2 similarities as \name, so that no method gains from a stronger encoder. QuoTA keeps its own scorer, a VLM judge that is part of the method itself. For methods that call a VLM more than once, every call counts toward their compute. Each baseline otherwise keeps the settings recommended by its authors and is implemented as follows.

\begin{itemize}\setlength{\itemsep}{0pt}\setlength{\parskip}{0pt}\setlength{\topsep}{2pt}
  \item \textbf{Uniform sampling}: evenly spaced frames, each with the same token count.
  \item \textbf{Top-K}: as described in the \mainpaper.
  \item \textbf{FOCUS}~\cite{zhu2026focus}: released code, with SigLIP~2 similarities in place of its BLIP scores.
  \item \textbf{Q-Frame}~\cite{zhang2025qframe}: released code, with SigLIP~2 similarities in place of its Long-CLIP scores and its three resolution tiers scaled to fill the token budget.
  \item \textbf{QuoTA}~\cite{luo2025quota}: released code, with its default prompts and its Qwen2-VL-2B judge.
  \item \textbf{AdaAlloc}~\cite{an2026adaalloc}: re-implemented from its paper, with SigLIP~2 similarities in place of its PE-Core scores. Because it makes three VLM calls per question, we scale the token budget of each call so that its total compute matches that of the other methods.
\end{itemize}

\section{Additional Results}
\label{sec:supp-results}

\begin{figure*}[t]
  \centering
  \includegraphics[width=\linewidth]{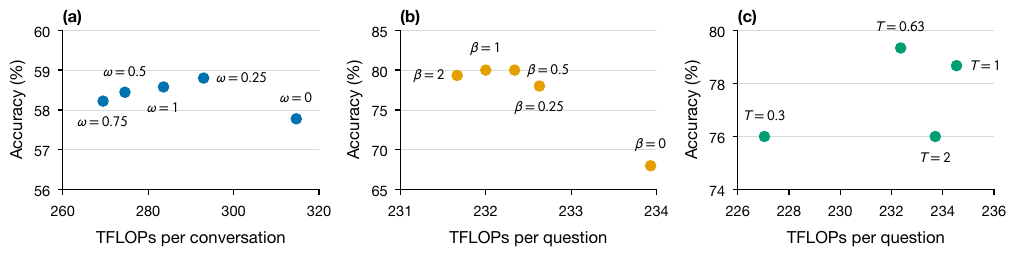}
  \caption{\textbf{Performance of \name under varying hyperparameters.} Accuracy against compute when varying (a) the attention weight $\omega$ on MT-Video-Bench, (b) the subtitle weight $\beta$ on Video-MME, and (c) the relevance temperature $T$ on Video-MME.}
  \label{fig:supp-sens}
\end{figure*}

\subsection{Per-Category Accuracy}
\label{sec:supp-category}

\Cref{fig:supp-radar} breaks down the accuracy in Tab.~1 of the \mainpaper{} by question category, where the 12 task types of Video-MME are grouped into six categories as in Q-Frame~\cite{zhang2025qframe}. Across the three benchmarks, \name achieves the highest accuracy in more categories than any baseline. On Video-MME, \name leads in five of the six categories with both VLMs, and its largest gains are on reasoning questions (4.2 and 6.8 points over the best baseline with Qwen3-VL-8B and InternVL3.5-4B, respectively). On MLVU, \name leads in six of the seven categories with Qwen3-VL-8B (one tied with Q-Frame) and in four with InternVL3.5-4B. The gains of \name thus hold across question categories and do not come from a single one.

\subsection{Solver Efficiency}
\label{sec:supp-solver}

\begin{table}[t]
  \centering
  \footnotesize
  \setlength{\tabcolsep}{2.6pt}
  \renewcommand{\arraystretch}{1.1}
  \begin{tabular}{l r r r r r r r}
    \toprule
    Frames & 30 & 60 & 120 & 240 & 420 & 1,800 & 3,600 \\
    \midrule
    \multicolumn{8}{l}{\textit{Solving time (s)}} \\
    Gurobi & 0.038 & 0.031 & 0.089 & 0.199 & 0.557 & 3.013 & 5.029 \\
    SCIP   & 1.943 & 3.989 & 9.655 & 23.927 & 51.662 & 600+ & 600+ \\
    CVXPY  & 0.010 & 0.013 & 0.039 & 0.040 & 0.080 & 0.763 & 1.794 \\
    Ours   & \textbf{0.006} & \textbf{0.007} & \textbf{0.009} & \textbf{0.031} & \textbf{0.025} & \textbf{0.132} & \textbf{0.256} \\
    \midrule
    \multicolumn{8}{l}{\textit{Objective}} \\
    Gurobi & 0.698 & 0.679 & 0.595 & 0.607 & 0.635 & 0.456 & 0.461 \\
    SCIP   & 0.698 & 0.679 & 0.595 & 0.607 & 0.635 & 0.000 & 0.083 \\
    CVXPY  & 0.698 & 0.679 & 0.596 & 0.607 & 0.636 & 0.457 & 0.461 \\
    Ours   & 0.698 & 0.679 & 0.595 & 0.607 & 0.635 & 0.456 & 0.461 \\
    \bottomrule
  \end{tabular}
  \caption{\textbf{\name solves the allocation problem quickly and near-optimally.} Solving time and objective value of iterative water-filling (Ours) and three off-the-shelf solvers.}
  \label{tab:supp-solver}
\end{table}

We compare iterative water-filling with three off-the-shelf solvers on synthetic videos of 30 to 3,600 candidate frames. Gurobi~\cite{gurobi} and SCIP~\cite{scip} are integer solvers that solve the discrete allocation problem to optimality. CVXPY~\cite{cvxpy} with the Clarabel solver~\cite{clarabel} solves the same continuous relaxation as iterative water-filling, so its objective value is an upper bound on the integer optimum. As shown in \cref{tab:supp-solver}, iterative water-filling is 1.3--7$\times$ faster than CVXPY and 4--23$\times$ faster than Gurobi. SCIP does not finish within 600\,s beyond 420 frames, and \cref{tab:supp-solver} reports the best solution it finds by then. Iterative water-filling also matches Gurobi's integer optimum to three decimals and stays within 0.001 of the upper bound.

\subsection{Sensitivity Analysis}
\label{sec:supp-sens}

We study how the accuracy and compute of \name change with three hyperparameters, namely the attention weight $\omega$, the subtitle weight $\beta$, and the relevance temperature $T$. \Cref{fig:supp-sens} varies each hyperparameter while keeping the others fixed. Each sweep runs on a subset of 500 questions. For the attention weight (\cref{fig:supp-sens}a), every $\omega>0$ improves over retrieval relevance alone ($\omega=0$) in both accuracy and compute, and the accuracies for $\omega>0$ differ by less than 1 point. For the subtitle weight (\cref{fig:supp-sens}b), the accuracies for $\beta>0$ stay within 2 points of each other. With $\beta=0$, no subtitle receives tokens and the prompt contains no subtitles at all, which reduces accuracy by 12 points. This setting differs from the ``w/o text allocation'' row of Tab.~2 of the \mainpaper, in which the VLM still receives the subtitles aligned with the selected frames, as the baselines do. For the relevance temperature (\cref{fig:supp-sens}c), accuracy is highest at the default $T=0.63$, remains within 1 point at $T=1$, and drops by about 3 points when $T$ is much smaller or larger. \name is therefore robust to $\omega$ and $\beta$ over a wide range, and to $T$ near its default value.

\end{document}